\documentclass{article}

\usepackage[preprint]{neurips_2026}

\usepackage[utf8]{inputenc}
\usepackage[T1]{fontenc}
\usepackage{url}
\usepackage{xcolor}
\usepackage{nicefrac}
\usepackage{microtype}

\usepackage{graphicx}
\usepackage{float}
\usepackage{adjustbox}
\usepackage{booktabs}
\usepackage{multirow}
\usepackage{colortbl}
\usepackage{amsmath}
\usepackage{amssymb}
\usepackage[accsupp]{axessibility}
\usepackage{subcaption}
\usepackage{wrapfig}
\usepackage[most]{tcolorbox}

\definecolor{linkblue}{rgb}{0.21,0.49,0.74}
\usepackage[breaklinks,colorlinks,allcolors=linkblue]{hyperref}
\usepackage[capitalize,noabbrev]{cleveref}

\makeatletter
\renewcommand{\@noticestring}{}
\makeatother

\begin{document}

\title{Reflective VLA: In-Context Action Consequences Make VLAs Generalize}

\author{%
  Qing Lian\textsuperscript{1}, Kent Yu\textsuperscript{1,2,3}, and Lei Zhang\textsuperscript{1,2,3} \\[4pt]
  \textsuperscript{1}\,Futian Laboratory \\
  \textsuperscript{2}\,International Digital Economy Academy (IDEA) \\
  \textsuperscript{3}\,Visincept \\[4pt]
  \texttt{lianqing1997@gmail.com}
}

\maketitle

\begin{abstract}
  Most vision-language-action (VLA) models are reactive: they predict the next action from the current instruction and observation, implicitly assuming that the current observation fully specifies the action-relevant state. In embodied control, however, embodiment-specific factors such as camera-to-robot geometry, robot calibration, or systematic actuation bias are often hard to identify from a single observation. As a result, reactive policies cannot reliably disambiguate these factors in general, overfitting to training environments and generalizing poorly at deployment. We propose Reflective VLA, which conditions each decision on a context of observation--action--consequence triplets. Each triplet records not only what the robot observed and executed, but also how the scene changed afterward, exposing the deployment-specific mapping from actions to observed effects. Architecturally, Reflective VLA routes all observation modalities through the VLM under shared attention, so the action expert reasons directly over past triplets and the current observation. A block-causal mask enables parallel multi-frame training without leakage and supports KV-cached real-time inference. On standard LIBERO and SimplerEnv-Bridge, Reflective VLA preserves strong in-distribution performance. Under distribution shift on LIBERO-Plus and the harder LIBERO-Plus-Hard, it improves average success rate by 5.4 and 4.2 percentage points over a matched reactive baseline.   Ablations with a matched history-only baseline further show that action consequences---rather than additional context length alone---are the key to cross-environment generalization. Project page: \url{https://lianqing11.github.io/reflective-vla-page/}.
\end{abstract}

\section{Introduction}\label{sec:intro}

Vision-language-action (VLA) systems build on pretrained vision-language backbones~\cite{qwen3-vl, qwen2.5-vl, paligemma} and fine-tune on large-scale robot demonstrations~\cite{oneill2024open, contributors2024agibotworldrepo} to produce language-conditioned control policies~\cite{zitkovich2023rt,kim2024openvla,black2024pi_0,liCogACTFoundationalVisionLanguageAction2024,nvidia2025gr00t}. By unifying scene understanding, instruction following, and low-level control in one model, they have substantially broadened the range of manipulation tasks a generalist policy can perform. Yet despite training on large and diverse datasets, current VLAs still generalize poorly to deployment environments unseen during training~\cite{fei2025libero-plus, zhang2025robustvla}.

\begin{figure}[t]
    \centering
    \includegraphics[width=\textwidth]{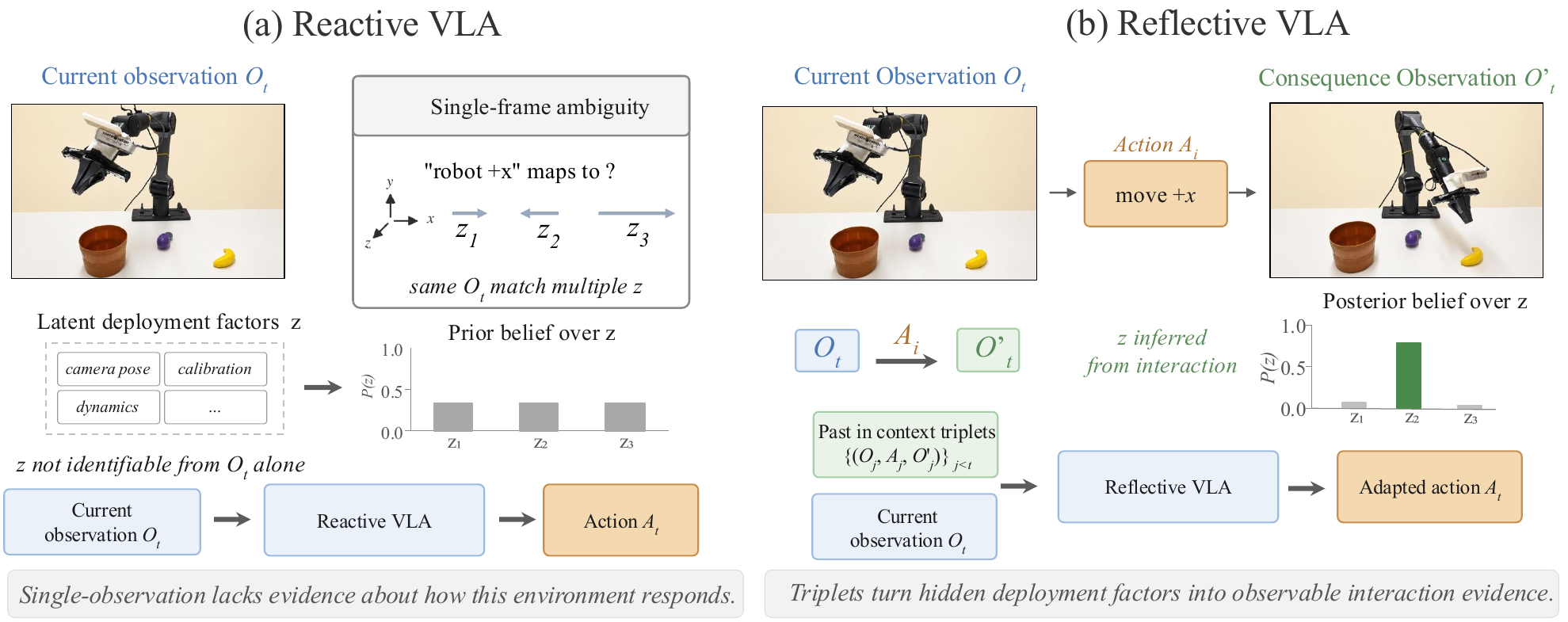}
    \caption{\textbf{From reactive to reflective control.}
    \textbf{(a) Reactive VLA.} Identifying embodiment-specific latent factors $z$---camera pose, calibration, etc.---from a single observation $O_t$ is ill-posed: many $z$ are consistent with the same frame, so $A_t$ overfits training embodiments.
    \textbf{(b) Reflective VLA.} Conditioning on past causal triplets $\{(O_j, A_j, O'_j)\}_{j<t}$, where $O'_j$ is the action-aligned observation after $A_j$, rules out deployments inconsistent with the evidence, sharpening $P(z\mid \text{context})$ and yielding an $A_t$ adapted to the current embodiment.}
    \vspace{-5mm}
    \label{fig:teaser}
\end{figure}

Cross-environment generalization in embodied control often requires inferring embodiment-specific factors that are hard to identify from single observations, such as camera geometry, robot calibration, and systematic actuation bias. In contrast, the \emph{interaction context}---an observation, an executed action, and the resulting observation---exposes these factors through how actions translate into observed changes: a commanded motion reveals camera extrinsics through its pixel displacement and calibration offsets through its pose residual. Existing VLAs condition on a single frame, omitting this evidence, so the policy must instead memorize the embodiments seen during training. Recent temporal-context and memory-based VLAs~\cite{rdt,memoryvla} improve state tracking, but lack explicit action--consequence binding, which is critical for identifying deployment-specific sensing and control factors.

We therefore cast cross-environment generalization in VLAs as an in-context learning (ICL) problem over causal interaction triplets. Given a small prompt of such triplets, the policy can infer the current deployment online from interaction feedback, analogous to how large language models infer task structure from few-shot demonstrations~\cite{brown2020language,xie2022an}. Instantiating this in a dual-system VLA raises two challenges. First, the prompt interleaves heterogeneous modalities---images, proprioception, and continuous action chunks---which must be packed into a causal sequence without bottlenecking the action decoder. Second, dual-system VLAs must propagate context through both the VLM prefix and the action expert; naive training repeats forward passes across target frames, while naive inference recomputes the history prefix at every step, making real-time control costly.

We address these challenges with \textbf{Reflective VLA}, an ICL framework for dual-system VLAs. The current observation is augmented with a small set of past triplets, each consisting of an observation, an executed action chunk, and the resulting observation. To pack heterogeneous modalities into a single causal sequence without bottlenecking the action decoder, all modalities share the VLM token space and a continuous action expert attends densely to the full prompt under shared attention. To make ICL training tractable, a block-causal mask supervises all $K$ context frames in a single forward pass instead of $K$ separate ones. The same causal structure supports KV-cached inference, so the extended context does not compromise real-time control at deployment.

We evaluate Reflective VLA on standard LIBERO and SimplerEnv, as well as on LIBERO-Plus~\cite{fei2025libero-plus} and our extension, LIBERO-Plus-Hard, which target deployment shifts in sensing, embodiment, and layout. Reflective VLA preserves strong standard-benchmark performance, improving over a matched reactive baseline by 0.7 and 5.3 points on LIBERO and SimplerEnv, respectively. Under deployment shifts, it improves held-out generalization by 5.4 and 4.2 points on LIBERO-Plus and LIBERO-Plus-Hard, respectively, without test-time fine-tuning. Ablations show that these gains come from observation--action--consequence evidence rather than longer context alone.

In summary, our main contributions are:
\begin{enumerate}
    \item We identify observation--action--consequence interaction context as a key signal for cross-environment VLA generalization, and formulate it as in-context learning over causal triplets.
    \item We introduce \textbf{Reflective VLA}, a dual-system VLA that places multimodal observations and historical action consequences in a shared VLM token space, with block-causal training and KV-cached real-time inference.
    \item Across LIBERO, SimplerEnv, LIBERO-Plus, and LIBERO-Plus-Hard, Reflective VLA improves standard-benchmark performance and held-out deployment generalization over matched reactive baselines; history-only ablations confirm that aligned consequence observations are the critical ingredient.
\end{enumerate}

\section{Related Work}\label{sec:related}

\noindent\textbf{Vision-language-action models.}
Recent generalist policies build on pretrained vision-language backbones and train on large-scale robot datasets to produce language-conditioned control~\cite{reed2022gato,brohan2023rt,zitkovich2023rt,kim2024openvla,team2024octo}. Architecturally, recent VLAs follow two main designs. One line casts control as autoregressive prediction over discretized or tokenized actions~\cite{zitkovich2023rt,kim2024openvla,kim2025fine,pertsch2025fast}. A more recent line decouples high-level reasoning from low-level control by pairing the VLM backbone with a dedicated continuous action expert based on diffusion or flow matching---a \emph{dual-system} design that preserves action fidelity at high control rates~\cite{black2024pi_0,liCogACTFoundationalVisionLanguageAction2024,lingbotvla,zhengXVLASoftPromptedTransformer2025a,nvidia2025gr00t}. We adopt this dual-system design and take an orthogonal direction: rather than scaling data or refining the action representation, we equip a fixed VLA with in-context interaction evidence as a new axis for cross-environment generalization,  without any test-time updates.

\noindent\textbf{Temporal context and memory for robot control.}
Temporal context is a natural response to partial observability in robot control. Imitation policies such as ACT, Diffusion Policy, and RDT use short observation histories and action chunks to stabilize visuomotor control~\cite{zhao2023learning,chi2023diffusion,rdt}; VLA systems such as RoboFlamingo, 4D-VLA, and MemoryVLA further incorporate visual history, spatiotemporal representations, or explicit memory for language-conditioned manipulation~\cite{roboflamingo,4dvla,memoryvla}. These methods improve state estimation, task progress tracking, and robustness to transient perceptual ambiguity. However, temporal history alone is not equivalent to interaction feedback: existing histories do not explicitly bind an executed action chunk to its aligned consequence. Reflective VLA therefore distinguishes temporal context from action-conditioned consequence context, and tests this distinction with history-only ablations that remove consequence observations from the same historical milestones.

\noindent\textbf{In-context adaptation for sequential decision making.}
In-context learning provides a complementary view of adaptation: a sequence model can infer the relevant task or environment from examples in its context~\cite{brown2020language,xie2022an}. This idea connects to meta-RL~\cite{duan2016rl2,finn2017maml,rakelly2019pearl} and to transformer policies that adapt behavior from contextual experience at test time, including Decision Transformer, Prompt-DT, and algorithm distillation~\cite{chen2021decision,promptdt,laskin2022incontext}.
Reflective VLA brings this viewpoint to embodied VLA control, but uses a different adaptation signal: structured multimodal observation--action--consequence triplets, without rewards, textual self-reflection, a separate system-identification module, or test-time parameter updates.

\section{Method}\label{sec:method}
\begin{figure}[t]
  \centering
  \includegraphics[width=0.95\textwidth]{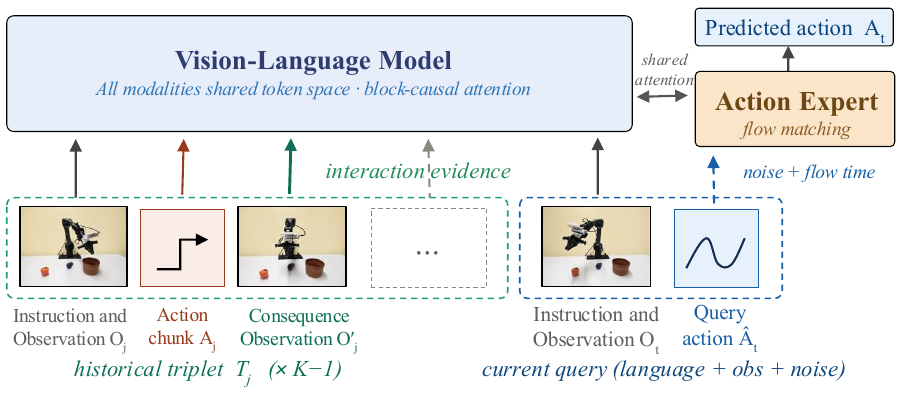}
  \vspace{-3mm}
  \caption{\textbf{Reflective VLA: observation--action--consequence in-context learning.} Past triplets $\{(O_j, A_j, O'_j)\}_{j=1}^{K-1}$ and the current observation $O_t$ share a single VLM token sequence; a flow-matching action expert attends to this prefix and denoises $\hat{A}_t$ into the predicted chunk $A_t$. The aligned consequence $O'_j$ carries the \emph{interaction evidence} that exposes environment factors.}
  \vspace{-3mm}
  \label{fig:method}
  \vspace{-3mm}
\end{figure}

Reflective VLA extends a reactive dual-system VLA with an in-context interface over observation--action--consequence triplets. We first define the reactive formulation, then describe how triplets are constructed and packed, how the model is trained on them in parallel using a block-causal mask, and how they are reused during online inference.

\subsection{Preliminary: Standard Reactive VLA Formulation}
\label{sec:prelim_reactive_vla}

At control step $t$, let $\mathcal{L}$ denote the language instruction, $\mathcal{O}_t$ the current multimodal observation, and $A_t=[a_t,\dots,a_{t+C-1}]$ an action chunk of horizon $C$. The observation may include third-person images, wrist images, and proprioceptive states. A standard reactive VLA predicts the next action chunk from only the current instruction and observation:
\begin{equation}
    A_t \sim \pi_\theta(\cdot \mid \mathcal{L}, \mathcal{O}_t).
\end{equation}
This interface covers recent dual-system VLAs that pair a VLM prefix with a continuous diffusion or flow-matching action expert. Although their action generation objectives differ, they share the same limitation for our purpose: the action expert conditions on the current observation, but not on past executed actions and their observed consequences.

\subsection{From Reactive to In-Context Generalization}
\label{sec:method_motivation}

A reactive VLA learns $\pi(A_t \mid \mathcal{L}, \mathcal{O}_t)$, predicting the next action from the current instruction and observation. Different deployments may share task semantics but differ in camera-to-robot geometry, robot calibration, or systematic actuation bias. We summarize these embodiment-specific factors as an environmental latent variable $z$, on which the appropriate action depends. Inferring $z$ from a single observation $\mathcal{O}_t$ is ill-posed: the mapping from $z$ to $\mathcal{O}_t$ is many-to-one, leaving the policy unable to disambiguate the current sensing and control conditions.

This missing evidence can be recovered by interaction feedback: when the robot executes an action and observes the resulting change, the observation--action--consequence relation reveals how the current deployment responds to commands. Formally, an adaptive policy can be written as a marginal over the posterior on $z$:
\begin{equation}
    \pi(A_t \mid \mathcal{L}, \mathcal{O}_t, \mathcal{H})
    =
    \int
    \pi(A_t \mid \mathcal{L}, \mathcal{O}_t, z)\,
    P(z \mid \mathcal{H})\, dz,
\end{equation}
where $\mathcal{H}$ is the interaction context accumulated so far. We do not explicitly estimate $z$ or compute this integral; the formulation serves only as a motivating abstraction, since transformers can implicitly approximate such posterior updates via in-context inference~\cite{xie2022an}.

The key requirement is that $\mathcal{H}$ is diagnostic of $z$. Under an idealized Markov view, if $\mathcal{H}$ consists of interaction triplets, the posterior factorizes as
\begin{equation}
    P(z \mid \mathcal{H})
    \propto
    P(z)
    \prod_i
    P_{\mathrm{env}}(\mathcal{O}'_i \mid \mathcal{O}_i, A_i, z),
\end{equation}
where $\mathcal{O}'_i$ is the observation after executing action chunk $A_i$. The likelihood depends explicitly on how the environment responds to executed actions, so a context containing only $(\mathcal{O}_i, A_i)$ pairs cannot expose the response term and provides limited evidence for identifying deployment-specific factors. 

Reflective VLA therefore conditions each decision on interaction context with explicit consequences:
\begin{equation}
    A_t \sim \pi_{\theta}\big( A_t \mid \mathcal{L}, \mathcal{H}, \mathcal{O}_t \big),
    \quad \mathcal{H} = \big\{(\mathcal{O}_i, A_i, \mathcal{O}'_i)\big\}_{i=1}^{K-1},
\end{equation}
where $\mathcal{H}$ is the observation--action--consequence context.

\subsection{Reflective VLA}
\label{sec:method_reflective}

\paragraph{Observation--action--consequence context.}
Reflective VLA represents recent interaction history as structured observation--action--consequence triplets. Each triplet stores the observation before an action chunk, the action chunk executed from that observation, and the resulting observation after the chunk completes. Since the policy predicts $C$-step action chunks, we define the consequence observation as $\mathcal{O}'_i = \mathcal{O}_{i+C}$ rather than the immediate next frame $\mathcal{O}_{i+1}$. This action-aligned consequence better captures the visible effect of the executed action, such as end-effector displacement and residual control error.

\begin{wrapfigure}{r}{0.5\linewidth}
  \vspace{-1.2em}
  \centering
  \includegraphics[width=\linewidth]{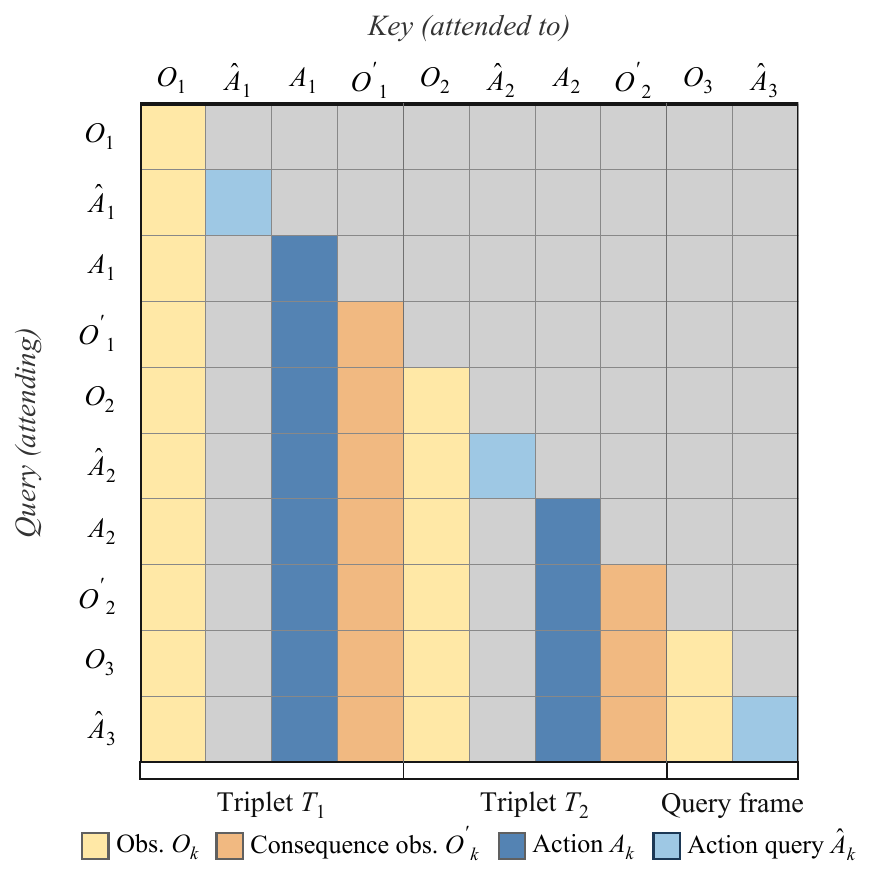}
  \vspace{-1.6em}
  \caption{\textbf{Block-causal mask.} Each query $\hat{A}_k$ attends to $\mathcal{L}$, prior triplets $T_{<k}$, and $\mathcal{O}_k$, while its own $A_k$, $\mathcal{O}'_k$, future triplets and queries are masked, supervising all targets in one forward.}

  \label{fig:masking}
  \vspace{-5.0em}
\end{wrapfigure}

At training time, $A_i$ is the demonstration action chunk and $\mathcal{O}'_i$ the observation after the chunk horizon; at deployment, $A_i$ is the chunk actually executed and $\mathcal{O}'_i$ the subsequent observation collected from the environment. Thus, each stored triplet represents a realized interaction rather than only a planned transition.
We embed each previous action chunk into eight tokens in the VLM token space via a learned projection $g_A$, and write the resulting prefix tokens simply as $A_i$.
The $i$-th context element is then $T_i = (\mathcal{L}, \tau_i, \mathcal{O}_i, A_i, \mathcal{O}'_i)$, where the language instruction $\mathcal{L}$ is re-inserted at the start of every triplet so that each unit is a self-contained $(\mathcal{L}, \mathcal{O}, A, \mathcal{O}')$ block, matching the per-triplet structure shown in \cref{fig:masking}.
We refer to these as causal triplets because the action token links a pre-action observation to its observed post-action consequence.

\paragraph{Multimodal prompt construction.}
Given historical milestones $m_1,\dots,m_{K-1}$ and the current query timestep $t$, Reflective VLA packs past triplets and the current observation---each prefixed by the language instruction $\mathcal{L}$---into a single multimodal sequence:
\begin{equation}
    X_{\mathrm{in}}
    =
    \Big[
    T_{m_1},\dots,T_{m_{K-1}},\;
    (\mathcal{L},\tau_t,\mathcal{O}_t)
    \Big].
\end{equation}
The historical triplets provide evidence about how the current deployment responds to actions, while the final $(\mathcal{L},\tau_t,\mathcal{O}_t)$ block serves as the query for predicting the next action chunk. The current target action is not inserted into the prefix; it is generated by the continuous action expert.

\paragraph{Shared-attention dual-system architecture.}
For the action decoder to use interaction context, all observation modalities must be visible to it. Some dual-system VLAs route only part of the observation through the VLM backbone or condition the action module on a compressed prefix~\cite{nvidia2025gr00t, zhengXVLASoftPromptedTransformer2025a}, which can bottleneck reasoning over (O, A, O') structure.
Reflective VLA therefore routes all observation modalities---third-person images, wrist images, and proprioception---into a shared VLM token sequence. Images are encoded by the visual backbone, while proprioceptive states and previous action chunks are projected into the token space with a learned non-linear head. A continuous flow-matching action expert is attached as a suffix that shares attention with the VLM prefix at every layer, following recent MoT-style VLA designs~\cite{liang2025mixtureoftransformers, black2024pi_0, lingbotvla}. The action expert can thus attend directly to prior observations, prior actions, observed consequences, temporal indices, and the current observation when generating $A_t$.

\paragraph{Block-causal training.}
In-context conditioning lengthens each training sequence by a factor of $K$, so supervising each context position with an independent forward pass would scale training cost as $\mathcal{O}(K)$. We instead borrow the packed-sequence idea from LM training and supervise all $K$ sampled frames jointly under a \emph{block-causal} attention mask, which lets every sampled frame act as both context for later targets and as a prediction target itself within a single forward pass. Because the same mask supervises positions with $0, 1, \ldots, K-1$ preceding triplets, training naturally covers the full range of context lengths the model will encounter at deployment.

\textit{Sampling and packing.}
We sample $K$ ordered frames $t_1 < \cdots < t_K$ from a trajectory and pack them into a single sequence: the first $K-1$ form the historical triplets and the $K$-th is the current query, but the block-causal mask supervises all $K$ frames jointly. As reflected in $T_i$, $\mathcal{L}$ is re-inserted at the start of every triplet (and the query block) so each unit is a self-contained $(\mathcal{L}, \mathcal{O}, A, \mathcal{O}')$ structure, and special tokens demarcate triplet boundaries so the action expert can unambiguously parse where each unit begins and ends. Because adjacent action chunks are temporally smooth, a fixed inter-frame stride invites the policy to extrapolate $A_{t_k}$ from past actions rather than learn from observed consequences; we therefore randomize the stride within a bounded range during training to break this shortcut.

\textit{Mask and objective.}
Each target frame $t_k$ is realized by a query slot $\hat{A}_{t_k}$ in the suffix flow-matching expert; \cref{fig:masking} visualizes the resulting attention pattern as a row in the mask. The query attends only to
\begin{equation}
    \mathcal{V}_{t_k} = \{T_{t_j}: j < k\} \cup \{(\mathcal{L}, \tau_{t_k}, \mathcal{O}_{t_k})\},
\end{equation}
i.e., all completed prior triplets (each carrying its own copy of $\mathcal{L}$) together with the current language-prefixed observation. It is masked from the prefix action token $A_{t_k}$ and consequence $\mathcal{O}'_{t_k}$ within its own triplet---both observed only after $A_{t_k}$ is executed---and from every subsequent triplet $\{T_{t_j}: j > k\}$; sibling query slots $\{\hat{A}_{t_j}\}_{j\neq k}$ are also mutually masked, so each prediction depends solely on prefix evidence. The training objective sums the flow-matching action loss over all valid targets:
\begin{equation}
    \mathcal{L}_{\mathrm{train}}=\sum_{k=1}^{K}\mathcal{L}_{\mathrm{act}}\big(A_{t_k};\mathcal{V}_{t_k}\big),
\end{equation}
where $\mathcal{L}_{\mathrm{act}}$ is the same flow-matching objective used by the base VLA. This yields dense multi-frame supervision in one forward pass while ensuring that each prediction is conditioned only on past completed interactions and the current observation. The first target $t_1$ has no preceding triplets and provides reactive supervision, while $t_2, \ldots, t_K$ provide ICL supervision with progressively longer context, as the staircase pattern along the diagonal of \cref{fig:masking} shows.

At deployment, Reflective VLA keeps a rolling buffer of the most recent $K{-}1$ triplets. Unlike training, where ground-truth action chunks populate the historical context, at inference each triplet stores the policy's own predicted chunk $\hat{A}_t$ together with the observation $\mathcal{O}_{t+C}$ actually reached after executing it---so the in-context prefix reflects the realized rollout rather than a teacher trajectory. At each step, only the new observation (or triplet, after execution) is encoded; VLM-side keys and values for past triplets are cached once and reused, keeping the per-step inference cost roughly constant.

\begin{table}[t]
  \centering
  \footnotesize
  \setlength{\tabcolsep}{3pt}
  \caption{\textbf{In-distribution evaluation on LIBERO and SimplerEnv-Bridge.}
  Success rates (\%) across the four standard LIBERO task suites and the SimplerEnv-Bridge benchmark; Avg denotes the unweighted mean over suites or tasks.
  $\dagger$ denotes our reproduced reactive baseline $\pi_{0.5}$.
  ``--'' indicates the result is not provided.}
  \label{tab:main_results}
  \begin{tabular}{l|ccccc|ccccc}
  \toprule
   & \multicolumn{5}{c|}{\textbf{LIBERO}} & \multicolumn{5}{c}{\textbf{SimplerEnv-Bridge}} \\
  \cmidrule(lr){2-6} \cmidrule(lr){7-11}
  Method & Spatial & Object & Goal & Long & Avg & Spoon & Carrot & Cube & Eggplant & Avg \\
  \midrule
  OpenVLA~\cite{kim2024openvla} & 84.7 & 88.4 & 79.2 & 53.7 & 75.9 & 4.2 & 0.0 & 8.3 & 45.8 & 14.6 \\
  CoT-VLA~\cite{zhao2025cotvla} & 87.5 & 91.6 & 87.6 & 69.0 & 81.1 & -- & -- & -- & -- & -- \\
  4D-VLA~\cite{4dvla} & 93.8 & 92.8 & 95.6 & 86.5 & 92.2 & -- & -- & -- & -- & -- \\
  ThinkAct~\cite{huang2025thinkact} & -- & -- & -- & -- & -- & 37.5 & 8.7 & 58.3 & 70.8 & 43.8 \\
  CogACT~\cite{liCogACTFoundationalVisionLanguageAction2024} & 97.2 & 98.0 & 90.2 & 88.8 & 93.2 & 71.7 & 50.8 & 15.0 & 67.5 & 51.3 \\
  InternVLA-M1~\cite{internvlam1} & 98.0 & \textbf{99.0} & 93.8 & 92.6 & 95.9 & 87.5 & 67.9 & 31.3 & \textbf{100.0} & 71.7 \\
  $\pi_0$~\cite{black2024pi_0} & 96.8 & 98.8 & 95.8 & 85.2 & 94.2 & -- & -- & -- & -- & -- \\
  MemoryVLA~\cite{memoryvla} & 98.4 & 98.4 & 96.4 & 93.4 & 96.5 & 75.0 & 75.0 & 37.5 & \textbf{100.0} & 71.9 \\
  GR00T N1.5~\cite{nvidia2025gr00t} & -- & -- & -- & -- & -- & 82.0 & 72.0 & 54.0 & 63.0 & 67.8 \\
  $\pi_{0.5}$~\cite{intelligence2025pi05} & \textbf{98.8} & 98.2 & 98.0 & 92.4 & 96.9 & -- & -- & -- & -- & -- \\
  \midrule
  Reactive baseline $\pi_{0.5}^{\dagger}$ & 97.5 & 98.2 & 97.8 & 94.0 & 96.9 & 91.7 & 79.2 & 70.8 & 50.0 & 72.9\\
  \textbf{Reflective VLA (ours)} & 98.4 & \textbf{99.0} & \textbf{98.2} & \textbf{94.6} & \textbf{97.6} & \textbf{95.8} & \textbf{83.3} & 79.2 & 54.2 & \textbf{78.2} \\
  \bottomrule
  \end{tabular}
  \vspace{-3mm}
\end{table}

\begin{table}[t]
  \centering
  \scriptsize
  \renewcommand{\arraystretch}{1.08}
  \setlength{\tabcolsep}{2.5pt}
  \caption{
    \textbf{Robustness under perturbations on LIBERO-Plus and LIBERO-Plus-Hard.}
    Success rates (\%) on LIBERO-Plus (7 standard perturbation categories) and
    LIBERO-Plus-Hard (2 additional harder shifts: Multi-camera shift
    (Camera$^{\dagger}$) and Robot calibration shift (Rob. Calib$^{\dagger}$)).
    }
  \label{tab:libero_plus}
  \begin{tabular}{l|ccccccc|c|cc|c}
  \toprule
  & \multicolumn{8}{c|}{\textbf{LIBERO-Plus}} & \multicolumn{3}{c}{\textbf{LIBERO-Plus-Hard}} \\
  \cmidrule(lr){2-9} \cmidrule(lr){10-12}
  Method & Camera & Robot & Lang & Light & Bg & Noise & Layout & Avg & Camera$^{\dagger}$ & Rob. Calib$^{\dagger}$ & Avg \\
  \midrule
  UniVLA~\cite{bu2025univla} & 1.8 & 46.2 & 69.6 & 69.0 & 81.0 & 21.2 & 31.9 & 45.8 & -- & -- & -- \\
  $\pi_{0}$~\cite{black2024pi_0} & 13.8 & 6.0 & 58.8 & 85.0 & 81.4 & 79.0 & 68.9 & 56.3 & -- & -- & -- \\
  OpenVLA-OFT~\cite{fei2025libero-plus} & 92.8 & 30.3 & 85.8 & 94.9 & 93.9 & 89.3 & \textbf{77.6} & 80.7 & 72.2 & 43.1 & 57.7 \\
  MemoryVLA~\cite{memoryvla} & 93.1 & 42.1 & 84.2 & 92.1 & 92.7 & 89.1 & 77.5 & 81.5 & 72.1 & 49.2 & 60.7 \\
  \midrule
  Reactive baseline $\pi_{0.5}^{\dagger}$ & 90.0 & 50.0 & 92.9 & 92.0 & 85.8 & 90.2 & 75.0 & 82.3 & 74.0 & 55.2 & 64.6 \\
  \textbf{Reflective VLA (ours)} & \textbf{95.7} & \textbf{72.9} & 92.2 & 92.1 & 92.7 & \textbf{95.4} & 73.0 & \textbf{87.7} & \textbf{76.3} & \textbf{61.3} & \textbf{68.8} \\
  \bottomrule
  \end{tabular}
  \vspace{-4mm}
\end{table}
\section{Experiments}\label{sec:experiments}
We evaluate Reflective VLA along three axes. First, we test whether adding interaction context preserves strong in-distribution performance on standard manipulation benchmarks. Second, we evaluate robustness under perturbations that change visual appearance, sensing, or embodiment-specific properties. Third, we ablate the designed components to illustrate the contribution of the observation--action--consequence structure rather than from longer context alone.

\subsection{Experimental Setup}
\label{sec:exp_setup}

\noindent\textbf{Simulation benchmarks.}
We evaluate on four simulation settings:
\textit{(i)}~LIBERO~\cite{liu2023libero}, the standard suite of 40 language-conditioned manipulation tasks across \textit{Spatial}, \textit{Object}, \textit{Goal}, and \textit{Long}, with a fixed camera and embodiment, used as our nominal in-distribution testbed;
\textit{(ii)}~SimplerEnv-Bridge~\cite{li2024simpler}, which evaluates VLAs trained on the BridgeDataV2~\cite{walke2023bridgedata} in the ManiSkill2~\cite{maniskill2} simulator, providing a real-to-sim transfer setting;
\textit{(iii)}~LIBERO-Plus~\cite{fei2025libero-plus}, which extends LIBERO with seven perturbation categories spanning sensing, language, and robot--environment configuration; and
\textit{(iv)}~LIBERO-Plus-Hard, our diagnostic extension targeting shifts in the action-to-observation mapping.

\noindent\textbf{A context-diagnostic benchmark.}
While LIBERO-Plus provides broad perturbation coverage, not all of its categories are equally diagnostic of interaction-conditioned adaptation: shifts in language, or background can largely be absorbed by invariances in the pretrained VLM backbones. We therefore introduce LIBERO-Plus-Hard with two perturbations designed so that single-frame evidence is insufficient and the action-to-observation mapping must be inferred from interaction:
\begin{itemize}
  \item \textbf{Multi-camera shift} (Camera$^{\dagger}$ in \cref{tab:libero_plus}). We jointly perturb the extrinsics of all camera views (third-person and wrist), so no single view preserves the nominal calibration; the camera-to-robot geometry must be recovered from the pixel-space displacement induced by past action chunks.
  \item \textbf{Robot calibration shift} (Rob. Calib$^{\dagger}$ in \cref{tab:libero_plus}). We inject an episode-level systematic offset between commanded and achieved end-effector motion, simulating calibration error, actuation bias, or mechanical backlash; this offset is unobservable from a static frame but recoverable from the residual between commanded actions and their consequences.
\end{itemize}

\begin{table}[t]
  \vspace{-2mm}
  \centering
  \footnotesize
  \setlength{\tabcolsep}{3pt}
  \renewcommand{\arraystretch}{1.08}
  \caption{\textbf{Ablations on the interaction context.}
  Success rates (\%) on Camera, Camera$^{\dagger}$, and Rob.\ Calib$^{\dagger}$; Avg denotes the mean.
  (a) varies what each context element contains, with $K$ fixed; (b) varies the number of context elements $K$ with the full $(O,A,O')$ structure.
  $K{=}1$ corresponds to the reactive baseline without historical triplets.}
  \label{tab:ablation}
  \vspace{-2mm}
  \begin{subtable}[t]{0.49\linewidth}
    \centering
    \caption{Context composition (fixed $K$).}
    \label{tab:ablation_context}
    \begin{adjustbox}{max width=\linewidth}
    \begin{tabular}{l|ccc|c}
      \toprule
      Context & Camera & Camera$^{\dagger}$ & Rob. Calib$^{\dagger}$ & Avg \\
      \midrule
      Reactive (no history) & 90.0 & 74.0 & 55.2 & 73.1\\
      $O$  & 88.8 & 73.2 & 55.0 & 72.3 \\
      $O,A$ & 89.1 & 74.9 & 57.4 & 73.8 \\
      $O,A,O'$ & \textbf{95.7} & \textbf{76.3} & \textbf{61.3} & \textbf{77.8} \\
      \bottomrule
    \end{tabular}
    \end{adjustbox}
  \end{subtable}
  \hfill
  \begin{subtable}[t]{0.49\linewidth}
    \centering
    \caption{Context length (full $(O,A,O')$).}
    \label{tab:ablation_length}
    \begin{adjustbox}{max width=\linewidth}
    \begin{tabular}{c|ccc|c}
      \toprule
      $K$ & Camera & Camera$^{\dagger}$ & Rob.\ Calib$^{\dagger}$ & Avg \\
      \midrule
      1 & 90.0 & 74.0 & 55.2 & 73.1 \\
      2 & 93.8 & 74.6 & 57.4 & 75.3 \\
      4 & 96.1 & 75.7 & 58.4 & 76.7 \\
      8 & \textbf{95.7} & \textbf{76.3} & \textbf{61.3} & \textbf{77.8} \\
      \bottomrule
    \end{tabular}
    \end{adjustbox}
  \end{subtable}
  \vspace{-4mm}
\end{table}

\noindent\textbf{Baselines.}
Our primary baseline, denoted $\pi_{0.5}^{\dagger}$, is a reproduced \emph{reactive} $\pi_{0.5}$ that uses the same backbone, training data, and  network parameters as Reflective VLA but with context length $K{=}1$, conditioning each prediction only on the current instruction and observation.
Where available and protocol-compatible, we additionally report published results for OpenVLA~\cite{kim2024openvla}, OpenVLA-OFT~\cite{kim2025fine}, UniVLA~\cite{bu2025univla}, CoT-VLA~\cite{zhao2025cotvla}, 4D-VLA~\cite{4dvla}, ThinkAct~\cite{huang2025thinkact}, CogACT~\cite{liCogACTFoundationalVisionLanguageAction2024}, InternVLA-M1~\cite{internvlam1}, $\pi_0$~\cite{black2024pi_0}, $\pi_{0.5}$~\cite{intelligence2025pi05}, MemoryVLA~\cite{memoryvla}, and GR00T~N1.5~\cite{nvidia2025gr00t}.

\noindent\textbf{Implementation details.}
Following $\pi_{0.5}$~\cite{intelligence2025pi05}, Reflective VLA adopts a Mixture-of-Transformers~\cite{liang2025mixtureoftransformers} dual-system design, pairing a pretrained VLM prefix with a flow-matching continuous action expert as the suffix under shared attention. Unless otherwise stated, we use action chunk size $C{=}10$, context length $K{=}8$, corresponding to seven historical triplets and one query observation, and bounded randomized milestone sampling. Additional architectural details, hyperparameters, and training settings are provided in \cref{app:method,app:training_details}.

\begin{wrapfigure}{r}{0.45\linewidth}
  \centering
  \includegraphics[width=\linewidth]{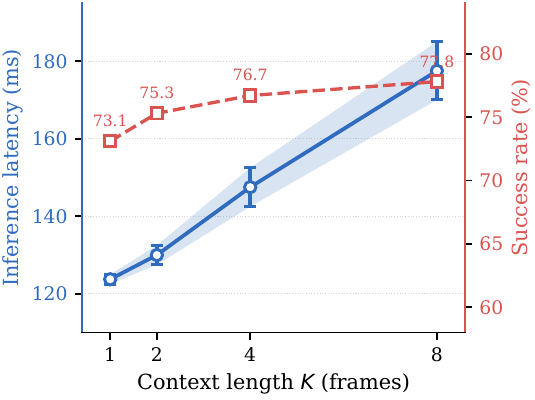}
  \vspace{-1.6em}
  \caption{Latency--accuracy trade-off across context length $K$ on the perturbation subset.}
  \vspace{-1.2em}
  \label{fig:latency_vs_k}
  \vspace{-0.5em}
\end{wrapfigure}

\subsection{Main Results}
\label{sec:exp_results}


\paragraph{In-distribution performance.}
\Cref{tab:main_results} shows that Reflective VLA preserves strong in-distribution performance on both LIBERO and SimplerEnv-Bridge. On LIBERO it reaches 97.6\% average success, achieving state-of-the-art---ahead of the published $\pi_{0.5}$ (96.9\%) and memory-based MemoryVLA (96.5\%)---and outperforming the matched reactive $\pi_{0.5}^{\dagger}$ baseline by 0.7 points. On SimplerEnv-Bridge, it reaches 78.2\% average success, again achieving state-of-the-art and improving over the reactive baseline by 5.3 points; since BridgeData~V2~\cite{walke2023bridgedata} spans diverse scenes, embodiments, and viewpoints, this gain reflects the model's ability to exploit in-context interaction evidence under embodiment-level variability already present in the training distribution. Performance on the \textit{Long} suite is also slightly improved (94.0\% $\to$ 94.6\%), indicating that interaction context does not hurt temporally extended manipulation. Together, these results show that adding structured interaction context does not compromise the base policy's nominal task-solving ability.

\paragraph{Robustness under perturbations.}
\Cref{tab:libero_plus} evaluates robustness on the seven standard LIBERO-Plus perturbation categories and two harder shifts in LIBERO-Plus-Hard; MemoryVLA is reproduced on the same training data as ours for a fair comparison. On LIBERO-Plus, Reflective VLA achieves 87.7\% average success, outperforming the matched reactive baseline (82.3\%), OpenVLA-OFT (80.7\%), and MemoryVLA (81.5\%; +6.2 pp). Reflective VLA improves on five of the seven categories, with the largest gains under \textit{Robot} (+22.9 pp), \textit{Background} (+6.9 pp), and \textit{Noise} (+5.2 pp); language perturbations are comparable between the two methods (92.9\% vs.\ 92.2\%), while \textit{Layout} is the main exception ($-$2.0 pp). The categories with the largest gains directly affect either the sensing interface or the robot's spatial configuration, both of which are difficult to identify from a single frame but exposed through how past actions translate into observed consequences.

\paragraph{Discussion.}
The two LIBERO-Plus-Hard shifts further stress-test this hypothesis. On this subset, Reflective VLA reaches 68.8\% average success, outperforming MemoryVLA (60.7\%; +8.1 pp) as well as the matched reactive baseline (64.6\%; +4.2 pp). Under \textit{Multi-camera shift} (Camera$^{\dagger}$), where the extrinsics of both third-person and wrist views are perturbed, Reflective VLA improves from 74.0\% to 76.3\%. Under \textit{Robot calibration shift} (Rob.\ Calib$^{\dagger}$), where the achieved end-effector motion deviates systematically from the commanded action, it improves from 55.2\% to 61.3\%. Unlike language or appearance perturbations, both shifts change the relationship between actions and their observed effects, making them difficult to identify from any single frame. The largest gain appears on Rob.\ Calib$^{\dagger}$, where the only diagnostic signal is the residual between commanded actions and their consequences---directly supporting the central design of Reflective VLA.

\paragraph{Context composition.}
\Cref{tab:ablation_context} isolates the effect of context composition with $K$ fixed. Adding observation history alone (\textsc{O}) yields no improvement over the reactive baseline (72.3\% vs.\ 73.1\%), and adding observation--action pairs without the resulting consequence (\textsc{O,A}) provides only a marginal gain (73.8\%). In contrast, the full observation--action--consequence context (\textsc{O,A,O$'$}) improves the average from 73.1\% to 77.8\%, a 4.7-point gain that is most pronounced on Rob.\ Calib$^{\dagger}$ (55.2\% $\to$ 61.3\%). This supports our hypothesis that the action-aligned consequence $\mathcal{O}'$ provides the key adaptation signal: temporal context alone, even when paired with the executed action, leaves the action-to-observation mapping unidentified.

\paragraph{Context length.}
\Cref{tab:ablation_length} studies the effect of context length. Increasing $K$ consistently improves robustness, from 73.1\% at $K{=}1$ to 75.3\% at $K{=}2$, 76.7\% at $K{=}4$, and 77.8\% at $K{=}8$. Most of the gain is captured by $K{=}4$ (+3.6 points over the reactive baseline), with diminishing returns thereafter, suggesting that a small number of recent interaction triplets already captures most of the useful deployment-specific evidence.

\paragraph{Latency--accuracy trade-off.}
\cref{fig:latency_vs_k} reports per-step latency against success rate during inference. Since past triplets are encoded once and reused, latency grows sub-linearly in $K$: $K{=}8$ is only $1.43{\times}$ slower than $K{=}1$ ($178$ vs.\ $124$\,ms) despite an $8{\times}$ longer prompt, while $K{=}4$ already recovers most of the accuracy gain at $1.19{\times}$ baseline latency---a practical operating point when tighter control rates are required. We cap $K$ at $8$ because longer contexts exceed our GPU memory budget during block-causal training; KV-cached inference itself can scale further.

\subsection{Real-World Experiments}
\label{sec:exp_realworld}

\begin{figure}[t]
  \centering
  \includegraphics[width=\linewidth]{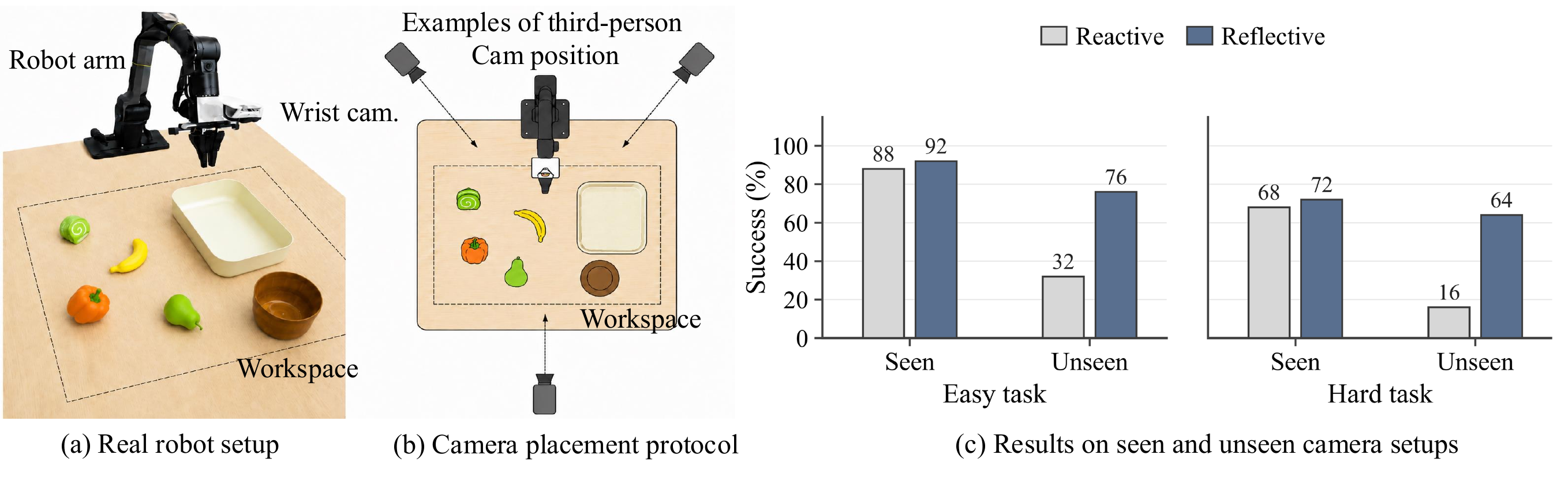}
  \caption{\textbf{Real-world setup.} (a) An Agilex Piper arm with RealSense D435i cameras over a tabletop workspace, with two tasks (place-into-box, place-into-bowl). (b) Third-person camera-placement protocol: ten placements span the left, front, and right of the workspace; demonstrations cover all ten, while evaluation uses five seen and five held-out placements drawn from the same regions.}
  \vspace{-3mm}
  \label{fig:realworld_setup}
  \vspace{-3mm}
\end{figure}

We further evaluate Reflective VLA under real-world cross-camera generalization. Following the protocol in \cref{fig:realworld_setup}, demonstrations cover ten third-person camera placements spanning the left, front, and right sides of the workspace; at test time we evaluate on five seen and five held-out placements ($N=25$ trials each) without test-time fine-tuning. We consider two tabletop tasks---placing an object into a large square box (primarily grasping) and into a small bowl (additionally requires camera-dependent placement). Full protocol details are in \cref{app:realworld}.

Reflective VLA preserves performance on seen viewpoints while substantially improving generalization to unseen ones. On seen placements, success improves only slightly over the reactive baseline (88\%$\to$92\% on box, 68\%$\to$72\% on bowl), indicating that interaction context does not compromise nominal performance when camera geometry is covered by training. Under unseen placements, the reactive baseline degrades sharply (32\% box, 16\% bowl) while Reflective VLA reaches 76\% and 64\%. Although the bowl task is harder overall due to its tighter placement tolerance, the gain on unseen viewpoints (+48\,pp) matches the box task (+44\,pp), suggesting that reflective context benefits both grasping robustness and camera-dependent spatial alignment.
\section{Conclusion}\label{sec:conclusion}

We cast cross-environment generalization in VLAs as in-context inference over causal triplets $(\mathcal{O}, A, \mathcal{O}')$, enabling adaptation from interaction feedback without test-time fine-tuning. Two findings support this view: a matched history-only ablation that omits the action aligned consequence $\mathcal{O}'$ recovers little of the gain, showing that context length alone is insufficient; and Reflective VLA closes a substantial portion of the OOD gap on the simulation and real world generalization environments. These gains come at constant model size---only a change in how the policy uses its context window. Scaling context length, extending reflection to longer horizons, and applying this framework to diverse embodiments are natural next steps.

\clearpage
{\small
\bibliographystyle{plainnat}
\bibliography{main}

@inproceedings{zitkovich2023rt,
  title={{RT-2}: Vision-Language-Action Models Transfer Web Knowledge to Robotic Control},
  author={Zitkovich, Brianna and Yu, Tianhe and Xu, Sichun and Xu, Peng and Xiao, Ted and Xia, Fei and Wu, Jialin and Wohlhart, Paul and Welker, Stefan and Wahid, Ayzaan and others},
  booktitle={CoRL},
  year={2023},
}

@inproceedings{kim2024openvla,
  title={{OpenVLA}: An Open-Source Vision-Language-Action Model},
  author={Kim, Moo Jin and Pertsch, Karl and Karamcheti, Siddharth and Xiao, Ted and Balakrishna, Ashwin and Nair, Suraj and Rafailov, Rafael and Foster, Ethan and Lam, Grace and Sanketi, Pannag and others},
  booktitle={CoRL},
  year={2024}
}

@inproceedings{team2024octo,
  title={{Octo}: An Open-Source Generalist Robot Policy},
  author={Team, Octo Model and Ghosh, Dibya and Walke, Homer and Pertsch, Karl and Black, Kevin and Mees, Oier and Dasari, Sudeep and Hejna, Joey and Kreiman, Tobias and Xu, Charles and others},
  booktitle={RSS},
  year={2024},
  doi={10.15607/RSS.2024.XX.090}
}

@inproceedings{black2024pi_0,
  title={$\pi_0$: A Vision-Language-Action Flow Model for General Robot Control},
  author={Black, Kevin and Brown, Noah and Driess, Danny and Esmail, Adnan and Equi, Michael and Finn, Chelsea and Fusai, Niccolo and Groom, Lachy and Hausman, Karol and Ichter, Brian and others},
  booktitle={RSS},
  year={2025},
  doi={10.15607/RSS.2025.XXI.010}
}

@inproceedings{kim2025fine,
  title={Fine-Tuning Vision-Language-Action Models: Optimizing Speed and Success},
  author={Kim, Moo Jin and Finn, Chelsea and Liang, Percy},
  booktitle={RSS},
  year={2025},
  doi={10.15607/RSS.2025.XXI.017}
}

@inproceedings{liu2023libero,
  title={{LIBERO}: Benchmarking Knowledge Transfer for Lifelong Robot Learning},
  author={Liu, Bo and Zhu, Yifeng and Gao, Chongkai and Feng, Yihao and Liu, Qiang and Zhu, Yuke and Stone, Peter},
  booktitle={NeurIPS},
  year={2023}
}

@article{pertsch2025fast,
  title={Fast: Efficient action tokenization for vision-language-action models},
  author={Pertsch, Karl and Stachowicz, Kyle and Ichter, Brian and Driess, Danny and Nair, Suraj and Vuong, Quan and Mees, Oier and Finn, Chelsea and Levine, Sergey},
  journal={arXiv preprint arXiv:2501.09747},
  year={2025}
}

@article{fei2025libero-plus,
  title={{LIBERO-Plus}: In-depth Robustness Analysis of Vision-Language-Action Models},
  author={Fei, Senyu and Wang, Siyin and Shi, Junhao and Dai, Zihao and Cai, Jikun and Qian, Pengfang and Ji, Li and He, Xinzhe and Zhang, Shiduo and Fei, Zhaoye and others},
  journal={arXiv preprint arXiv:2510.13626},
  year={2025}
}

@article{lingbotvla,
  title={A Pragmatic VLA Foundation Model},
  author={Wu, Wei and Lu, Fan and Wang, Yunnan and Yang, Shuai and Liu, Shi and Wang, Fangjing and Zhu, Qian and Sun, He and Wang, Yong and Ma, Shuailei and others},
  journal={arXiv preprint arXiv:2601.18692},
  year={2026}
}

@article{liang2025mixtureoftransformers,
  title={Mixture-of-Transformers: A Sparse and Scalable Architecture for Multi-Modal Foundation Models},
  author={Weixin Liang and Lili Yu and Liang Luo and Srini Iyer and Ning Dong and Chunting Zhou and Gargi Ghosh and Mike Lewis and Wen-tau Yih and Luke Zettlemoyer and Xi Victoria Lin},
  journal={Transactions on Machine Learning Research},
  year={2025},
}

@inproceedings{xie2022an,
  title={An Explanation of In-Context Learning as Implicit Bayesian Inference},
  author={Sang Michael Xie and Aditi Raghunathan and Percy Liang and Tengyu Ma},
  booktitle={ICLR},
  year={2022},
}

@inproceedings{brown2020language,
  title={Language Models are Few-Shot Learners},
  author={Tom B. Brown and Benjamin Mann and Nick Ryder and Melanie Subbiah and Jared Kaplan and Prafulla Dhariwal and Arvind Neelakantan and Pranav Shyam and Girish Sastry and Amanda Askell and Sandhini Agarwal and Ariel Herbert-Voss and Gretchen Krueger and Tom Henighan and Rewon Child and Aditya Ramesh and Daniel M. Ziegler and Jeffrey Wu and Clemens Winter and Chris Hesse and Mark Chen and Eric Sigler and Mateusz Litwin and Scott Gray and Benjamin Chess and Jack Clark and Christopher Berner and Sam McCandlish and Alec Radford and Ilya Sutskever and Dario Amodei},
  booktitle={NeurIPS},
  volume={33},
  year={2020},
}

@inproceedings{promptdt,
  title={Prompting Decision Transformer for Few-Shot Policy Generalization},
  author={Xu, Mengdi and Shen, Yikang and Zhang, Shun and Lu, Yuchen and Zhao, Ding and Tenenbaum, Joshua B. and Gan, Chuang},
  booktitle={ICML},
  year={2022}
}

@inproceedings{laskin2022incontext,
  title={In-context Reinforcement Learning with Algorithm Distillation},
  author={Laskin, Michael and Wang, Luyu and Oh, Junhyuk and Parisotto, Emilio and Spencer, Stephen and Steigerwald, Richie and Strouse, DJ and Hansen, Steven and Filos, Angelos and Brooks, Ethan and Gazeau, Maxime and Sahni, Himanshu and Singh, Satinder and Mnih, Volodymyr},
  booktitle={ICLR},
  year={2023}
}

@inproceedings{roboflamingo,
  title={Vision-Language Foundation Models as Effective Robot Imitators},
  author={Li, Xinghang and Liu, Minghuan and Zhang, Hanbo and Yu, Cunjun and Xu, Jie and Wu, Hongtao and Cheang, Chilam and Jing, Ya and Zhang, Weinan and Liu, Huaping and Li, Hang and Kong, Tao},
  booktitle={ICLR},
  year={2024}
}

@article{liCogACTFoundationalVisionLanguageAction2024,
  title={CogACT: A Foundational Vision-Language-Action Model for Synergizing Cognition and Action in Robotic Manipulation},
  author={Li, Qixiu and Liang, Yaobo and Wang, Zeyu and Luo, Lin and Chen, Xi and Liao, Mozheng and Wei, Fangyun and Deng, Yu and Xu, Sicheng and Zhang, Yizhong and Wang, Xiaofan and Liu, Bei and Fu, Jianlong and Bao, Jianmin and Chen, Dong and Shi, Yuanchun and Yang, Jiaolong and Guo, Baining},
  journal={arXiv preprint arXiv:2411.19650},
  year={2024}
}

@inproceedings{memoryvla,
  title={MemoryVLA: Perceptual-Cognitive Memory in Vision-Language-Action Models for Robotic Manipulation},
  author={Shi, Hao and Xie, Bin and Liu, Yingfei and Sun, Lin and Liu, Fengrong and Wang, Tiancai and Zhou, Erjin and Fan, Haoqiang and Zhang, Xiangyu and Huang, Gao},
  booktitle={ICLR},
  year={2026}
}

@inproceedings{zhao2025cotvla,
  title={CoT-VLA: Visual Chain-of-Thought Reasoning for Vision-Language-Action Models},
  author={Zhao, Qingqing and Lu, Yao and Kim, Moo Jin and Fu, Zipeng and Zhang, Zhuoyang and Wu, Yecheng and Li, Zhaoshuo and Ma, Qianli and Han, Song and Finn, Chelsea and Handa, Ankur and Lin, Tsung-Yi and Wetzstein, Gordon and Liu, Ming-Yu and Xiang, Donglai},
  booktitle={CVPR},
  pages={1702--1713},
  year={2025},
}

@inproceedings{4dvla,
  title={4D-VLA: Spatiotemporal Vision-Language-Action Pretraining with Cross-Scene Calibration},
  author={Zhang, Jiahui and Chen, Yurui and Xu, Yueming and Huang, Ze and Zhou, Yanpeng and Yuan, Yu-Jie and Cai, Xinyue and Huang, Guowei and Quan, Xingyue and Xu, Hang and Zhang, Li},
  booktitle={NeurIPS},
  year={2025}
}

@inproceedings{zhengXVLASoftPromptedTransformer2025a,
  title={X-VLA: Soft-Prompted Transformer as Scalable Cross-Embodiment Vision-Language-Action Model},
  author={Zheng, Jinliang and Li, Jianxiong and Wang, Zhihao and Liu, Dongxiu and Kang, Xirui and Feng, Yuchun and Zheng, Yinan and Zou, Jiayin and Chen, Yilun and Zeng, Jia and Zhang, Ya-Qin and Pang, Jiangmiao and Liu, Jingjing and Wang, Tai and Zhan, Xianyuan},
  booktitle={ICLR},
  year={2026}
}

@article{nvidia2025gr00t,
  title={{GR00T N1}: An Open Foundation Model for Generalist Humanoid Robots},
  author={{NVIDIA} and Bjorck, Johan and Casta{\~n}eda, Fernando and Cherniadev, Nikita and Da, Xingye and Ding, Runyu and Fan, Linxi and others},
  journal={arXiv preprint arXiv:2503.14734},
  year={2025}
}

@inproceedings{chi2023diffusion,
  title={Diffusion Policy: Visuomotor Policy Learning via Action Diffusion},
  author={Chi, Cheng and Xu, Zhenjia and Feng, Siyuan and Cousineau, Eric and Du, Yilun and Burchfiel, Benjamin and Tedrake, Russ and Song, Shuran},
  booktitle={RSS},
  year={2023},
  doi={10.15607/RSS.2023.XIX.026}
}

@inproceedings{rdt,
  title={RDT-1B: A Diffusion Foundation Model for Bimanual Manipulation},
  author={Liu, Songming and Wu, Lingxuan and Li, Bangguo and Tan, Hengkai and Chen, Huayu and Wang, Zhengyi and Xu, Ke and Su, Hang and Zhu, Jun},
  booktitle={ICLR},
  year={2025}
}

@article{reed2022gato,
  title={A Generalist Agent},
  author={Reed, Scott and Zolna, Konrad and Parisotto, Emilio and Colmenarejo, Sergio G{\'o}mez and Novikov, Alexander and Barth-Maron, Gabriel and Gim{\'e}nez, Mai and Sulsky, Yury and Kay, Jackie and Springenberg, Jost Tobias and others},
  journal={Transactions on Machine Learning Research},
  year={2022}
}

@inproceedings{brohan2023rt,
  title={RT-1: Robotics Transformer for Real-World Control at Scale},
  author={Brohan, Anthony and Brown, Noah and Carbajal, Justice and Chebotar, Yevgen and Dabis, Joseph and Finn, Chelsea and Gopalakrishnan, Keerthana and Hausman, Karol and Herzog, Alexander and Hsu, Jasmine and others},
  booktitle={RSS},
  year={2023},
  doi={10.15607/RSS.2023.XIX.025}
}

@inproceedings{oneill2024open,
  title={Open X-Embodiment: Robotic Learning Datasets and RT-X Models},
  author={{Open X-Embodiment Collaboration} and O'Neill, Abby and Rehman, Abdul and Maddukuri, Abhiram and Gupta, Abhishek and Padalkar, Abhishek and Lee, Abraham and Pooley, Acorn and Gupta, Agrim and Mandlekar, Ajay and others},
  booktitle={ICRA},
  year={2024},
}

@inproceedings{chen2021decision,
  title={Decision Transformer: Reinforcement Learning via Sequence Modeling},
  author={Chen, Lili and Lu, Kevin and Rajeswaran, Aravind and Lee, Kimin and Grover, Aditya and Laskin, Michael and Abbeel, Pieter and Srinivas, Aravind and Mordatch, Igor},
  booktitle={NeurIPS},
  year={2021}
}

@inproceedings{zhao2023learning,
  title={Learning Fine-Grained Bimanual Manipulation with Low-Cost Hardware},
  author={Zhao, Tony Z. and Kumar, Vikash and Levine, Sergey and Finn, Chelsea},
  booktitle={RSS},
  year={2023},
  doi={10.15607/RSS.2023.XIX.016}
}

@article{duan2016rl2,
  title={RL$^2$: Fast Reinforcement Learning via Slow Reinforcement Learning},
  author={Duan, Yan and Schulman, John and Chen, Xi and Bartlett, Peter L. and Sutskever, Ilya and Abbeel, Pieter},
  journal={arXiv preprint arXiv:1611.02779},
  year={2016}
}

@inproceedings{finn2017maml,
  title={Model-Agnostic Meta-Learning for Fast Adaptation of Deep Networks},
  author={Finn, Chelsea and Abbeel, Pieter and Levine, Sergey},
  booktitle={ICML},
  year={2017}
}

@inproceedings{rakelly2019pearl,
  title={Efficient Off-Policy Meta-Reinforcement Learning via Probabilistic Context Variables},
  author={Rakelly, Kate and Zhou, Aurick and Quillen, Deirdre and Finn, Chelsea and Levine, Sergey},
  booktitle={ICML},
  year={2019}
}

@inproceedings{bu2025univla,
  title={UniVLA: Learning to Act Anywhere with Task-centric Latent Actions},
  author={Bu, Qingwen and Yang, Yanting and Cai, Jisong and Gao, Shenyuan and Ren, Guanghui and Yao, Maoqing and Luo, Ping and Li, Hongyang},
  booktitle={RSS},
  year={2025},
  doi={10.15607/RSS.2025.XXI.014}
}

@article{qwen3-vl,
  title={Qwen3-VL Technical Report},
  author={Bai, Shuai and Cai, Yuxuan and Chen, Ruizhe and Chen, Keqin and Chen, Xionghui and Cheng, Zesen and Deng, Lianghao and Ding, Wei and Gao, Chang and Ge, Chunjiang and Ge, Wenbin and Guo, Zhifang and Huang, Qidong and Huang, Jie and Huang, Fei and Hui, Binyuan and Jiang, Shutong and Li, Zhaohai and Li, Mingsheng and Li, Mei and Li, Kaixin and Lin, Zicheng and Lin, Junyang and Liu, Xuejing and Liu, Jiawei and Liu, Chenglong and Liu, Yang and Liu, Dayiheng and Liu, Shixuan and Lu, Dunjie and Luo, Ruilin and Lv, Chenxu and Men, Rui and Meng, Lingchen and Ren, Xuancheng and Ren, Xingzhang and Song, Sibo and Sun, Yuchong and Tang, Jun and Tu, Jianhong and Wan, Jianqiang and Wang, Peng and Wang, Pengfei and Wang, Qiuyue and Wang, Yuxuan and Xie, Tianbao and Xu, Yiheng and Xu, Haiyang and Xu, Jin and Yang, Zhibo and Yang, Mingkun and Yang, Jianxin and Yang, An and Yu, Bowen and Zhang, Fei and Zhang, Hang and Zhang, Xi and Zheng, Bo and Zhong, Humen and Zhou, Jingren and Zhou, Fan and Zhou, Jing and Zhu, Yuanzhi and Zhu, Ke},
  journal={arXiv preprint arXiv:2511.21631},
  year={2025}
}

@article{qwen2.5-vl,
  title={Qwen2.5-VL Technical Report},
  author={Bai, Shuai and Chen, Keqin and Liu, Xuejing and Wang, Jialin and Ge, Wenbin and Song, Sibo and Dang, Kai and Wang, Peng and Wang, Shijie and Tang, Jun and Zhong, Humen and Zhu, Yuanzhi and Yang, Mingkun and Li, Zhaohai and Wan, Jianqiang and Wang, Pengfei and Ding, Wei and Fu, Zheren and Xu, Yiheng and Ye, Jiabo and Zhang, Xi and Xie, Tianbao and Cheng, Zesen and Zhang, Hang and Yang, Zhibo and Xu, Haiyang and Lin, Junyang},
  journal={arXiv preprint arXiv:2502.13923},
  year={2025}
}

@article{paligemma,
  title={{PaliGemma}: A versatile {3B VLM} for transfer},
  author={Beyer, Lucas and Steiner, Andreas and Pinto, Andr{\'e} Susano and Kolesnikov, Alexander and Wang, Xiao and Salz, Daniel and Neumann, Maxim and Alabdulmohsin, Ibrahim and Tschannen, Michael and Bugliarello, Emanuele and Unterthiner, Thomas and Keysers, Daniel and Koppula, Skanda and Liu, Fangyu and Grycner, Adam and Gritsenko, Alexey and Houlsby, Neil and Kumar, Manoj and Rong, Keran and Eisenschlos, Julian and Kabra, Rishabh and Bauer, Matthias and Bo{\v{s}}njak, Matko and Chen, Xi and Minderer, Matthias and Voigtlaender, Paul and Bica, Ioana and Balazevic, Ivana and Puigcerver, Joan and Papalampidi, Pinelopi and Henaff, Olivier and Xiong, Xi and Soricut, Radu and Harmsen, Jeremiah and Zhai, Xiaohua},
  journal={arXiv preprint arXiv:2407.07726},
  year={2024}
}

@article{contributors2024agibotworldrepo,
  title={{AgiBot World Colosseo}: A Large-scale Manipulation Platform for Scalable and Intelligent Embodied Systems},
  author={{AgiBot-World-Contributors} and Bu, Qingwen and Cai, Jisong and Chen, Li and Cui, Xiuqi and Ding, Yan and Feng, Siyuan and Gao, Shenyuan and He, Xindong and others},
  journal={arXiv preprint arXiv:2503.06669},
  year={2025}
}

@article{zhang2025robustvla,
  title={RobustVLA: Robustness-Aware Reinforcement Post-Training for Vision-Language-Action Models},
  author={Zhang, Hongyin and Zhang, Shuo and Jin, Junxi and Zeng, Qixin and Li, Runze and Wang, Donglin},
  journal={arXiv preprint arXiv:2511.01331},
  year={2025}
}

@article{li2024simpler,
  title={Evaluating Real-World Robot Manipulation Policies in Simulation},
  author={Li, Xuanlin and Hsu, Kyle and Gu, Jiayuan and Pertsch, Karl and Mees, Oier and Walke, Homer Rich and Fu, Chuyuan and Lunawat, Ishikaa and Sieh, Isabel and Kirmani, Sean and Levine, Sergey and Wu, Jiajun and Finn, Chelsea and Su, Hao and Vuong, Quan and Xiao, Ted},
  journal={arXiv preprint arXiv:2405.05941},
  year={2024}
}

@inproceedings{maniskill2,
  title={{ManiSkill2}: A Unified Benchmark for Generalizable Manipulation Skills},
  author={Gu, Jiayuan and Xiang, Fanbo and Li, Xuanlin and Ling, Zhaoyuan and Liu, Xiqiang and Mu, Tongzhou and Tang, Yihe and Tao, Stone and Wei, Xinyue and Yao, Yuzhe and Yuan, Xiaodi and Xie, Pengwei and Huang, Zhiao and Chen, Rui and Su, Hao},
  booktitle={International Conference on Learning Representations},
  year={2023}
}

@inproceedings{walke2023bridgedata,
  title={{BridgeData V2}: A Dataset for Robot Learning at Scale},
  author={Walke, Homer Rich and Black, Kevin and Zhao, Tony Z. and Vuong, Quan and Zheng, Chongyi and Hansen-Estruch, Philippe and He, Andre Wang and Myers, Vivek and Kim, Moo Jin and Du, Max and Lee, Abraham and Fang, Kuan and Finn, Chelsea and Levine, Sergey},
  booktitle={Conference on Robot Learning},
  pages={1723--1736},
  volume={229},
  year={2023}
}

@article{intelligence2025pi05,
  title={{$\pi_{0.5}$}: a Vision-Language-Action Model with Open-World Generalization},
  author={{Physical Intelligence}},
  journal={arXiv preprint arXiv:2504.16054},
  year={2025}
}

@inproceedings{huang2025thinkact,
  title={{ThinkAct}: Vision-Language-Action Reasoning via Reinforced Visual Latent Planning},
  author={Huang, Chi-Pin and Wu, Yueh-Hua and Chen, Min-Hung and Wang, Yu-Chiang Frank and Yang, Fu-En},
  booktitle={NeurIPS},
  year={2025}
}

@article{internvlam1,
  title={{InternVLA-M1}: A Spatially Guided Vision-Language-Action Framework for Generalist Robot Policy},
  author={{Intern Robotics}},
  journal={arXiv preprint arXiv:2510.13778},
  year={2025}
}
}

\clearpage
\appendix

\section*{Appendix}
\addcontentsline{toc}{section}{Appendix}

This appendix provides the implementation and experimental details needed to reproduce Reflective VLA.
\Cref{app:method} describes the architecture components, triplet construction, context-buffer behavior, block-causal masking, and flow-matching inference procedure.
\Cref{app:liberohard} specifies the two diagnostic perturbation families used in LIBERO-Plus-Hard.
\Cref{app:exp} details the training data construction, optimization settings, and evaluation protocol for LIBERO, LIBERO-Plus, LIBERO-Plus-Hard, and SimplerEnv-Bridge.
\Cref{app:realworld} documents the real-world robot setup, control interface, tasks, and camera-placement protocol.
\Cref{app:repro_limitations} summarizes reproducibility resources, external assets, and current limitations.

\section{Implementation Details}\label{app:method}
\subsection{Architecture components}

Reflective VLA uses the same model family and training budget as the matched reactive baseline, and changes only the conditioning interface.
We implement it with a dual-system Mixture-of-Transformers (MoT) architecture: a pretrained VLM encodes the language and multimodal observation prefix, and a continuous flow-matching transformer action expert is attached as a suffix with shared self-attention to the prefix.
We instantiate the VLM with either PaliGemma-3B~\cite{paligemma} or Qwen3-VL-2B~\cite{qwen3-vl}, and use an action expert with hidden dimension 1024.
Third-person and wrist images are processed through the native VLM visual pathway, while proprioceptive states and historical action chunks are projected into the token space by lightweight two-layer nonlinear fully connected projectors.
Each historical action chunk is represented by eight learned tokens.
The matched reactive baseline uses the same backbone, projectors, action expert, optimizer, and data, but sets $K{=}1$ so that no historical triplets are provided.

\subsection{Triplet construction and context buffer}\label{app:triplet_buffer}

For each control step $t$, the query input contains the instruction, the current multimodal observation, and a bounded history of completed interaction triplets. The current observation comprises the third-person view, left/right wrist views (when available), and the proprioceptive state. For a chunk horizon $C$, each historical triplet is stored as
\begin{equation}
  T_i = \big(\tau_i, \mathcal{O}_i, A_i, \mathcal{O}_{i+C}\big),
\end{equation}
where $A_i=[a_i,\ldots,a_{i+C-1}]$ is the executed action chunk, projected by $g_A$ into eight learned action tokens. We use $C{=}10$ for LIBERO and SimplerEnv-Bridge, and $C{=}30$ for the real-world experiments.

During training, triplets are sampled only from completed past interactions of the same episode, so each consequence $\mathcal{O}_{i+C}$ strictly precedes the query. To prevent the model from extrapolating the target action from a fixed temporal pattern of neighboring actions, we add a randomized stride of $[0,15]$ environment steps to the backward spacing.

At inference, the policy maintains a rolling FIFO buffer of completed triplets from the current episode: after each executed chunk, the newly formed triplet is appended and the oldest is evicted once the buffer is full. Triplets are drawn from the buffer with a fixed backward stride for deterministic input layout, and at the start of an episode the model simply conditions on whatever context is available.

\subsection{Prompt packing details}
\begin{tcolorbox}[
  colback=gray!4,
  colframe=gray!35,
  boxrule=0.4pt,
  arc=1pt,
  left=4pt,
  right=4pt,
  top=4pt,
  bottom=4pt,
  title={VLM input template},
  fonttitle=\bfseries\footnotesize,
  fontupper=\ttfamily\footnotesize
]
<|im\_start|>user \{instruction\} \\[2pt]
\textcolor{gray}{\# historical triplets} \\
<|frame\_start|> <Third-Person OBS$_i$> <Wrist OBS$_i$> <PROPRIO$_i$> \\
\quad <|action\_start|><ACTION$_i$><|action\_end|> \\
\quad <Third-Person OBS$_{i+C}$> <Wrist OBS$_{i+C}$> <PROPRIO$_{i+C}$> <|frame\_end|> \\
<|frame\_sep|> \quad $\cdots$ \quad <|frame\_sep|> \\
\textcolor{gray}{\# current query, no target action or consequence} \\
<|frame\_start|> <Third-Person OBS$_t$> <Wrist OBS$_t$> <PROPRIO$_t$> \\
\end{tcolorbox}


\section{LIBERO-Plus-Hard Specification}\label{app:liberohard}

\subsection{Perturbation overview}

LIBERO-Plus-Hard extends LIBERO-Plus with two diagnostic perturbation families targeting factors that are hard to identify from a single frame but directly govern the action-to-observation mapping: camera geometry and robot calibration. We denote them Camera$^\dagger$ and Rob.\ Calib$^\dagger$ in the result tables. Task semantics, objects, and language instructions are unchanged.

For each rollout, the perturbation parameters are sampled once at the start of the episode and held fixed throughout. All methods share the same task initializations, seeds, and sampled perturbation instances, and the success criterion follows the original LIBERO task-completion signal. The latent perturbation is not exposed to the model but can be inferred from observation--action--consequence triplets collected during the current episode.

\subsection{Multi-camera shift}

The multi-camera shift perturbs the visual sensing interface while leaving task, object layout, and robot dynamics unchanged. For the third-person view, we sample an episode-level camera transform with azimuth in $[-75^\circ, 75^\circ]$, elevation in $[0^\circ, 15^\circ]$, distance scale in $[1.0, 2.0]$, and endpoint rotations in $[-10^\circ, 10^\circ]$. For the wrist view, we sample a field of view in $[60^\circ, 90^\circ]$ and one of four image transforms (identity, horizontal flip, vertical flip, $180^\circ$ rotation), applied consistently throughout the rollout. Both views remain visually valid but their geometry differs from the nominal LIBERO setting, so the action-to-pixel mapping must be re-identified online.

\subsection{Robot calibration shift}

The robot calibration shift perturbs the mapping from commanded actions to the motion executed by the simulator, simulating systematic calibration error, actuation bias, or mechanical offset. At the start of each rollout, we sample a fixed calibration bias and apply it to every absolute end-effector command before stepping the environment. Under the default moderate setting, the translational bias is sampled independently along each Cartesian axis from $[-10, 10]$\,mm, and the rotational bias is a fixed axis-angle offset with magnitude up to $3^\circ$; the gripper command is unchanged. We only collect the replayed trajectories that successfully complete the task.

The bias is unobservable from any single frame but recoverable from completed triplets, since the residual between the commanded action stored in context and the realized next observation directly reveals it.

\section{Experimental Protocol}\label{app:exp}
\subsection{Training details}\label{app:training_details}

\paragraph{Training data construction.}
For the standard LIBERO and SimplerEnv-Bridge experiments, we use the official LIBERO demonstrations and BridgeData V2 trajectories, with action targets converted to absolute end-effector control and a chunk horizon of $C{=}10$.

For LIBERO-Plus, we use the released LIBERO-Plus training set and convert the trajectories to absolute end-effector control via replay.
For LIBERO-Plus-Hard, we follow the same replay-based generation pipeline on the original LIBERO demonstrations, additionally injecting our two diagnostic perturbations during replay: camera-pose shifts and commanded-action biases.
We merge the two sets and train a single model on the union, yielding 43{,}546 trajectories in total.

\paragraph{Optimization.}
All models are trained with AdamW ($\beta_1{=}0.9$, $\beta_2{=}0.95$, weight decay $0.01$, gradient clipping at norm $1.0$) under bf16 mixed precision with DeepSpeed ZeRO-2.
The action expert and projectors use a peak learning rate of $10^{-4}$; the VLM uses a $0.1\times$ multiplier and is held frozen for the initial period in \cref{tab:supp_training_hparams}, after which it follows the same linear warm-up and cosine decay schedule as the rest of the model.

For Reflective VLA on LIBERO/LIBERO-Plus we use $K{=}8$ context frames, and $K{=}4$ on SimplerEnv-Bridge. Adjacent context frames are separated by one action-chunk stride plus a random gap, sampled from $[0,15]$ environment steps for LIBERO-family experiments and $[0,5]$ for SimplerEnv-Bridge; at evaluation the gap is set to zero for deterministic fixed-stride selection.
\begin{table}[t]
  \centering
  \footnotesize
  \setlength{\tabcolsep}{4pt}
  \caption{Training hyperparameters. Batch size is reported per GPU; all reported runs use 8 H20 GPUs.}
  \label{tab:supp_training_hparams}
  \begin{tabular}{l|c|c|c}
    \toprule
    Setting & Steps & Batch/GPU & $K$ \\
    \midrule
    LIBERO reactive baseline           & 90k  & 32 & 1 \\
    LIBERO Reflective VLA              & 50k  & 4  & 8 \\
    LIBERO-Plus / Hard reactive        & 50k  & 32 & 1 \\
    LIBERO-Plus / Hard Reflective VLA  & 90k  & 4  & 8 \\
    SimplerEnv-Bridge reactive         & 80k  & 32 & 1 \\
    SimplerEnv-Bridge Reflective VLA   & 160k & 6  & 4 \\
    \bottomrule
  \end{tabular}
\end{table}

\subsection{Evaluation protocol}\label{app:eval_protocol}
All models are evaluated from a fixed checkpoint without test-time fine-tuning, using the same pipeline for the reactive baseline and Reflective VLA. For Reflective VLA, the context buffer is reset per episode and populated online from the policy's own executed chunks and reached observations.

\paragraph{LIBERO.} We evaluate the four official suites (\textit{Spatial}, \textit{Object}, \textit{Goal}, \textit{Long}) under absolute end-effector control, with 50 rollouts per task (seed 42) and a horizon of 800 steps (900 for \textit{Long}). Success follows the environment's task-completion signal; we report per-suite success and the unweighted average.

\paragraph{LIBERO-Plus and LIBERO-Plus-Hard.} For LIBERO-Plus we evaluate the seven standard perturbation categories with one rollout per task; for LIBERO-Plus-Hard we additionally evaluate the multi-camera shift and the robot-calibration shift (default moderate level) on the LIBERO base suites. Perturbations are generated deterministically from task and rollout ids, so all methods see identical instances.

\paragraph{SimplerEnv-Bridge.} We evaluate the four WidowX tasks (spoon on towel, carrot on plate, cube stacking, eggplant in basket) with 24 rollouts per task, reporting per-task success and the unweighted average.

All tables report success rates in percent, averaged over the reported suites or perturbation categories.

\subsection{More qualitative results}
\begin{figure}[H]
  \centering
  \includegraphics[width=\linewidth]{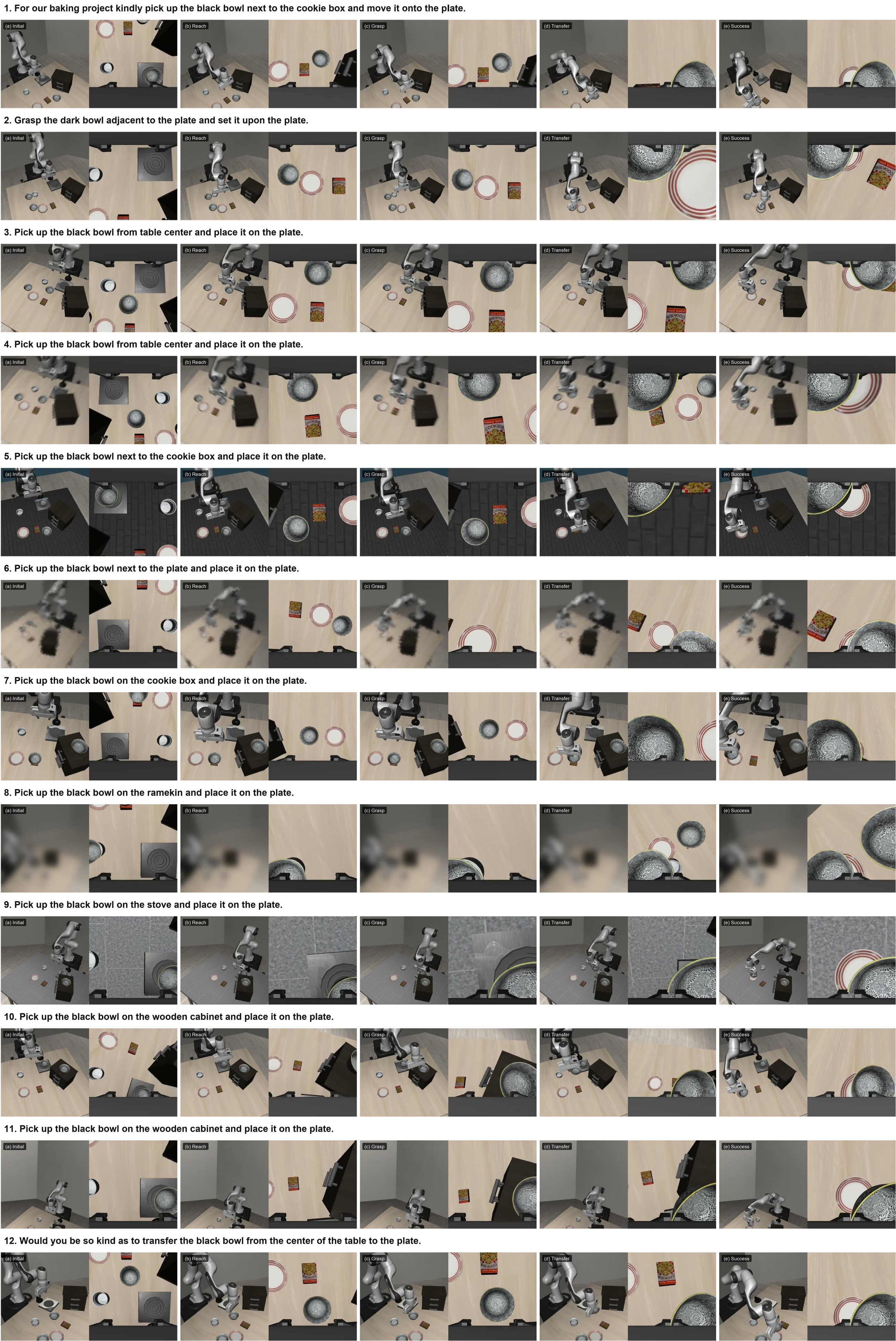}
  \caption{Qualitative results of reflective VLA on LIBERO-Plus-Hard dataset.}
  \label{fig:qualitative_overview}
\end{figure}

In \cref{fig:qualitative_overview}, we show the qualitative results of Reflective VLA on LIBERO-Plus-Hard dataset.

\section{Real-World Protocol}\label{app:realworld}

\subsection{Hardware and control interface}

We use a tabletop setup with an Agilex Piper arm and Intel RealSense D435i cameras. Each policy receives the language instruction, the current third-person view, and the proprioceptive state, and predicts chunked delta end-effector pose and gripper commands with horizon $C{=}30$, executed through the same interface for the reactive baseline and Reflective VLA. For Reflective VLA, the context buffer is reset per trial and populated online from executed chunks and their resulting observations; the reactive baseline conditions only on the current observation and instruction. All hardware, controller, camera streams, and task initializations are identical across methods, with no test-time fine-tuning or camera-specific calibration.
\subsection{Tasks and camera placements}

We use two language-conditioned pick-and-place tasks: a \textit{box} task (placement into a large square box) and a \textit{bowl} task (placement into a smaller bowl, requiring tighter spatial alignment). We collect 500 demonstrations across ten third-person camera placements spanning the left, front, and right sides of the workspace. Evaluation uses both seen and held-out placements from the same regions, preserving task semantics and hardware while changing camera-to-robot geometry. For each task and condition, we run five trials per placement ($N=25$ total) with matched initial object configurations; success is judged by a human (target object inside the container at rollout end).

\paragraph{Statistical significance.}
We report Wilson 95\% confidence intervals in \cref{sec:exp_realworld}. On the \textit{box} task, the reactive baseline reaches 88\% [70.0, 95.8] (seen) and 32\% [17.2, 51.6] (held-out), versus 92\% [75.0, 97.8] and 76\% [56.6, 88.5] for Reflective VLA. On the \textit{bowl} task, the corresponding numbers are 68\% [48.4, 82.8] / 16\% [6.4, 34.7] for the baseline and 72\% [52.4, 85.7] / 64\% [44.5, 79.8] for Reflective VLA. On both tasks, the held-out intervals do not overlap, indicating that the cross-camera generalization gain is significant at the 95\% level; seen-placement intervals overlap, consistent with the small in-distribution gap.
\section{Reproducibility, Assets, and Limitations}\label{app:repro_limitations}

\paragraph{Assets.}
Our simulated experiments build on public assets from LIBERO, LIBERO-Plus, BridgeData V2, and SimplerEnv. Pretrained vision-language backbones are obtained from their official releases under their respective licenses. The LIBERO-Plus-Hard perturbations are specified procedurally in our evaluation scripts (\cref{app:liberohard}). Real-world experiments use the hardware described in \cref{app:realworld}.

\paragraph{Limitations and Future work.}
Reflective VLA predicts the first chunk reactively, and assumes consequences are observable within the chunk horizon---delayed effects or contact-rich dynamics may need longer contexts. Three further constraints stem from our compute and data budget: (i) we cap context at $K{=}8$ frames due to training memory; (ii) providing history can induce a shortcut where the policy extrapolates from past action chunks instead of reasoning from the current observation, partially mitigated by frame-boundary tokens and stride randomization; (iii) in-context generalization benefits from data diversity, so scaling training data should yield further gains. Our real-world study is also limited to tabletop cross-camera generalization with few trials per condition.
\newpage


\end{document}